\documentclass{article}

\usepackage{graphicx} 

\usepackage{multirow} 

\usepackage{algpseudocode} 
\usepackage{algorithm}
\usepackage{algorithmicx}
\usepackage{mathtools}
\usepackage{nccmath}
\usepackage{multirow}
\usepackage{bigdelim}
\usepackage{color, colortbl}

\usepackage{amsfonts}       
\usepackage{nicefrac}       
\usepackage{microtype}      
\usepackage{xcolor}         
\usepackage{graphicx}
\usepackage{amssymb}
\usepackage{float}
\usepackage{amsmath}
\usepackage{amsthm}
\usepackage{wrapfig}
\usepackage{todonotes}
\usepackage{subcaption}
\usepackage{bbm}

\usepackage{graphicx}
\usepackage{caption}

\newcommand{\inputnum}{3} 

\newcommand{\hiddennum}{5}  

\newcommand{\outputnum}{2}

\usepackage{subcaption}

\usepackage{blindtext}
\usepackage{geometry}
\usepackage{bbold} 

\usepackage{graphicx} 

\usepackage{amsmath} 

\usepackage{xcolor} 

\usepackage{mathtools}

\usepackage{hyperref} 
\usepackage{amssymb} 

\usepackage[backend=biber, maxnames=99]{biblatex} 

\usepackage{xcolor}
\hypersetup{
    colorlinks,
    linkcolor={red!50!black},
    citecolor={blue!50!black},
    urlcolor={blue!80!black}
}

\usepackage{float}

\newtheorem{proposition}{Proposition}

\usepackage{mathtools}

\usepackage{amsmath,amssymb}

\usepackage{tikz}
\usepackage{pgfplots}

\newcommand{\Prob}{\mathbb{P}}

\title{Confident magnitude-based neural network pruning}
\author{Joaquin Alvarez \footnote{ITAM}}
\date{}

\begin{document}
\maketitle
\begin{abstract}
Pruning neural networks has proven to be a successful approach to increase the efficiency and reduce the memory storage of deep learning models without compromising performance. Previous literature has shown that it is possible to achieve a sizable reduction in the number of parameters of a deep neural network without deteriorating its predictive capacity in one-shot pruning regimes. Our work builds beyond this background in order to provide rigorous uncertainty quantification for pruning neural networks reliably, which has not been addressed to a great extent in previous literature focusing on pruning methods in computer vision settings. We leverage recent techniques on distribution-free uncertainty quantification to provide finite-sample statistical guarantees to compress deep neural networks, while maintaining high performance. Moreover, this work presents experiments in computer vision tasks to illustrate how uncertainty-aware pruning is a useful approach to deploy sparse neural networks safely.






\end{abstract}

\maketitle

\section{Introduction}









The recent progress in the capabilities of deep neural networks and their high performance in complex tasks has been accompanied by an increase in their size. This represents a challenge because computations become more expensive and memory requirements become more  demanding to store such large models. Pruning deep neural networks has shown promising solutions to these problems \cite{blalock2020state}.

Neural network pruning refers to a broad technique that reduces a neural network's size, for example by inducing sparsity in its parameters, with the hope of conserving a satisfying level of performance. In this way, the pruned model is a compressed version of the original full neural network. With such compression, the sparse model guarantees efficient computation and low memory requirements compared to its original dense architecture. A vast literature supports the benefits of pruning \cite{han2015learning,zhu2017prune}.

Critically, however, pruning neural networks involves a tradeoff which should be addressed carefully. Despite the remarkable strengths of pruning highlighted by some recent literature, we should also point out that pruning a neural network is a delicate method which should be implemented accounting for uncertainty in the models' performance degradation. This is because when some weights of a neural network are zeroed this may induce harmful properties in the pruned model, even if the pruning is restricted to a small ratio of low-magnitude weights or seemingly inconsequential parameters. As an example of this phenomena in the celebrated context of LLMs, consider \cite{yin2024pruning}. This motivates us to consider the relevance of distribution-free uncertainty quantification to control and preserve the quality of compressed predictive algorithms with high confidence. 

In this paper we present a way to bridge the best of both worlds: using deep learning models with pruning while offering rigorous finite-sample statistical guarantees on their performance. This allows to obtain the high performance levels offered by large sized neural networks using low computational resources through pruning while accounting for uncertainty in the pruning calibration. 

Put from another perspective, the classical statistical toolbox does not apply when dealing with deep neural networks, especially if they are treated as black-boxes. For example, there is no such thing a t-test for the significance of every parameter of a neural network, as it may have thousands or even millions of parameters with unknown properties. Confident neural network pruning allows to some extent to assess the importance of the weights of a neural network without the need to sacrifice formal statistical arguments.

It is worth to emphasize why we find this work to be relevant in the current state-of-the-art deep network pruning methods from the literature. Importantly, there are diverse pruning approaches. Regardless of the diversity in the frameworks, a critical element in almost any pruning algorithm is a stop criterion that determines \emph{when to stop pruning} \cite{vadera2021methods, blalock2020state, diao2023pruning} in those methods that begin with a fully connected neural network and sparsify gradually. Some of the most common stopping conditions are to do a validation, which terminates the pruning when the sparse model's accuracy falls below certain threshold or stop pruning if the validation accuracy repeatedly decreases. Indeed, this involves doing \emph{multiple} tests implicitly, because after pruning the user checks the performance of the pruned model and decides whether to keep pruning or not. However, to our knowledge, uncertainty quantification in stop criteria and calibration have not yet been considered crucial in the fast-evolving pruning literature as meta-analysis and survey research in the field show \cite{blalock2020state, cheng2023surveydeepneuralnetwork, vadera2021methods}, with \cite{zhao2022pruning,laufer2023efficiently} being among the few exceptions that we have found. In this work we address \emph{stop criterion with uncertainty quantification} taking multiple comparisons into account, focusing on computer vision problems. Our work offers a rigorous approach to stop criteria leveraging recent frameworks on distribution-free uncertainty quantification \cite{bates2021distributionfree,angelopoulos2022learn} and adapts beyond accuracy to other stopping metrics of potential interest. We tackle the use of formal statistical methods rather than having to make overly conservative corrections to terminate the pruning process.

Consider a calibration dataset $\mathcal{D}_{\text{cal}} \coloneqq\{(X_i,Y_i)\}_{i=1}^{n}$ corresponding to a random sample from a distribution $\mathbb{P}_{XY}$ and a pretrained neural network $f:\mathcal{X}\longrightarrow[0,1]^{M\times N}$. The neural network is used to make predictions which we denote by $\hat{Y}_{\text{full}}(X)$. Consider a pruning ratio (also referred to as the sparsification hyper-parameter) $\lambda\in [0,1)$ which zeroes $\lambda\times 100\%$ of the weights with the smallest magnitude in the neural network. We denote by $f_{\lambda}$ the sparse version of $f$ and the corresponding point prediction is denoted by $\hat{Y}_{\lambda}(X)$. As a warm-up example, in the MNIST dataset we work with $M=1$ and $N=10$, which corresponds to the $10$ digits on each image, and we can make point predictions with $\hat{Y}_{\text{full}}(X)=\texttt{argmax}(f(X))\in \{0,1,\dots, 9\}$. We evaluate the predictive performance of the compressed model with respect to the fully connected model through a loss function $\ell (\hat{Y}_{\text{full}}(X), \hat{Y}_{\lambda}(X))$. Let $\alpha\in (0,1)$ be a performance degradation tolerance and $\delta\in (0,1)$. Our approach provides a pruning stopping criterion satisfying:

$$\mathbb{P}\big(\mathbb{E}[\ell (\hat{Y}_{\text{full}}(X), \hat{Y}_{\lambda}(X))]\leq \alpha \big)\geq 1-\delta,$$

where the probability is computed on the calibration dataset which is used to obtain the largest possible value of $\lambda$ controlling the risk which we define as $R(\lambda)\coloneqq\mathbb{E}[\ell (\hat{Y}_{\text{full}}(X), \hat{Y}_{\lambda}(X))]$. This formulation might be beneficial in the case where the dataset is not labeled, which means that there is no need to have the corresponding $Y_i$ for each $X_i$ ,$i\in \{1,\dots, n\}$, or it may be of interest in the case when the comparison is restricted to the full neural network relative to the pruned neural network. But we also consider losses that use the $(X,Y)$ pairs and compare the pruned models' predictions to the true response variables, namely formulations satisfying
$$\mathbb{P}\big(\mathbb{E}[\ell(Y, \hat{Y}_{\lambda}(X))]\leq \alpha \big)\geq 1-\delta,$$

where the calibration set is required to contain the true labels to evaluate comparing the predictions to the true response variables. We also consider cases where there may be more than one hyper-parameter involved in the calibration.

\section{Related work}

When it comes to the broad spectrum of pruning techniques, we should highlight that pruning can take place at different stages. There exist pruning methods before training \cite{lee2019snipsingleshotnetworkpruning}, during training \cite{zhu2017prune}, or after training \cite{han2015learning}. Moreover, pruning might be done in a \emph{three-stage pipeline} of training, pruning and fine-tuning a deep neural network \cite{han2015learning} with many iterations on the loop. Interestingly, that approach is inspired by the learning process of the mammalian brain \cite{han2015learning}. Or it can also be implemented in a one-shot regime \cite{benbaki2023fastchitaneuralnetwork}, without running iterations of pruning and fine-tuning or retraining the pruned model. Iterative pruning may achieve greater sparsity but requires more computation \cite{cheng2023surveydeepneuralnetwork}. The achieved sparsity of pruning methods can be structured or unstructured \cite{liu2019rethinking, yin2024pruning, blalock2020state}. In this work we will focus on pruning pretrained neural networks in a one-shot regime, but the ideas may be adapted to other settings. 

What makes pruning a non-trivial problem is that it typically entails a tradeoff between efficiency (in terms of computational costs and memory) and performance: more pruning usually involves more efficiency and less memory use but reduces the performance \cite{blalock2020state}. The immediate question is how many weights to remove accounting for this tradeoff?  Our approach can offer an answer to such question by providing a rigorous criteria to determine how many weights can be zeroed when pruning a neural network based on a user-specified loss degradation tolerance on a calibration dataset.

This work is related to \cite{zhao2022pruning}, with important differences in its methods. Their work focuses on conformal methods taking the size of the prediction sets as the main calibration concern (the so-called inefficiency of predictions as the authors name it), whereas we work on general loss functions using methods with similar marginal statistical guarantees but leveraging different techniques. We also consider the MNIST dataset \cite{lecun1998gradient} in our experiments, but we consider different setups and different applications as theirs.

We should also highlight \cite{laufer2023efficiently} and \cite{laufer2023risk}, which consider the Learn then Test framework in accuracy-cost tradeoffs, and specifically pruning large-scale Transformer models in a natural language processing (NLP) setting. While their pruning approach also leverages distribution-free uncertainty quantification to prune neural networks, their work considered different pruning techniques than in this work. They consider token pruning, early exiting and head pruning. Their main focus is on their proposed three dimensional transformer pruning scheme with experiments primarily centered on text classification, whereas this work lies within the unstructured global magnitude-based pruning setting, and is mainly concerned with computer vision experiments.

This work is inspired by recent success of distribution-free uncertainty quantification methods \cite{angelopoulos2022learn, bates2021distributionfree} to accelerate text generation in language models \cite{schuster2022confident} with statistical guarantees on the quality of the adaptive language model as well as in the calibration of early stopping rules in LLMs \cite{ringel2024early} and early-exit approaches with risk control \cite{jazbec2024fast}, all of which lie within the bigger scope of efficient and trustworthy AI deployment using conformal prediction and related frameworks. Our work is related to \cite{schuster2022confident} in that we use multiple testing algorithms and the main calibration hyper-parameter in this work holds similar properties to theirs. The role of the calibration hyper-parameter makes the predictions more efficient in our work as well as in theirs, and there is an inherent tradeoff because larger values of the calibration hyper-parameter lead to higher performance degradation using  approximate monotonic loss functions which reflect the inverse relationship between cost computation and predictive performance. A similar approach is considered in \cite{ringel2024early} where earlier stopping leads to faster classifications but higher risks in general.



\section{Calibrating the pruning ratio}
In this section we introduce the setup of our approach which we will use for our experimental examples in computer vision.

Pruning methods might take different perspectives on the information offered by the weights of a pretrained model. Essentially, the value of the weights represents what the model was able to learn about the structure and the properties of the training dataset with respect to the training loss function. There are several approaches to determine the importance these weights through score functions. For example, the magnitude of the weights may be of interest, but also first order information, which is relevant to study the sensitivity of the loss to the weights of the network and the Hessian of the loss function evaluated on these weights may contain relevant information too \cite{NIPS1989_6c9882bb}.

\begin{figure}[H]
     \centering
     \begin{subfigure}[b]{0.4\textwidth}
         \centering
        \begin{tikzpicture}[scale=1.1]
\foreach \i in {1,...,\inputnum}
{
    \node[circle, 
        minimum size = 6mm,
        fill=orange!60] (Input-\i) at (0,-\i) {};
}
 
\foreach \i in {1,...,\hiddennum}
{
    \node[circle, 
        minimum size = 6mm,
        fill=teal!50,
        yshift=(\hiddennum-\inputnum)*5 mm
    ] (Hidden-\i) at (2.5,-\i) {};
}
 
\foreach \i in {1,...,\outputnum}
{
    \node[circle, 
        minimum size = 6mm,
        fill=blue!40,
        yshift=(\outputnum-\inputnum)*5 mm
    ] (Output-\i) at (5,-\i) {};
}

\draw[->,draw=gray,  line width=.2mm] (Input-1) -- (Hidden-1);
\draw[->,draw=gray,  line width=.9mm] (Input-1) -- (Hidden-2);
\draw[->,draw=gray,  line width=.1mm] (Input-1) -- (Hidden-3);
\draw[->,draw=gray,  line width=1mm] (Input-1) -- (Hidden-4);
\draw[->,draw=gray,  line width=.7mm] (Input-1) -- (Hidden-5);

\draw[->,draw=gray,  line width=.9mm] (Input-2) -- (Hidden-1);
\draw[->,draw=gray,  line width=.3mm] (Input-2) -- (Hidden-2);
\draw[->,draw=gray,  line width=1mm] (Input-2) -- (Hidden-3);
\draw[->,draw=gray,  line width=.5mm] (Input-2) -- (Hidden-4);
\draw[->,draw=gray,  line width=.7mm] (Input-2) -- (Hidden-5);

\draw[->,draw=gray,  line width=.1mm] (Input-3) -- (Hidden-1);
\draw[->,draw=gray,  line width=.1mm] (Input-3) -- (Hidden-2);
\draw[->,draw=gray,  line width=.4mm] (Input-3) -- (Hidden-3);
\draw[->,draw=gray,  line width=.3mm] (Input-3) -- (Hidden-4);
\draw[->,draw=gray,  line width=.9mm] (Input-3) -- (Hidden-5);

\draw[->,draw=gray,  line width=.1mm] (Hidden-1) -- (Output-1);
\draw[->,draw=gray,  line width=.5mm] (Hidden-1) -- (Output-2);

\draw[->,draw=gray,  line width=.6mm] (Hidden-2) -- (Output-1);
\draw[->,draw=gray,  line width=.1mm] (Hidden-2) -- (Output-2);

\draw[->,draw=gray,  line width=.3mm] (Hidden-3) -- (Output-1);
\draw[->,draw=gray,  line width=.2mm] (Hidden-3) -- (Output-2);

\draw[->,draw=gray,  line width=.9mm] (Hidden-4) -- (Output-1);
\draw[->,draw=gray,  line width=.5mm] (Hidden-4) -- (Output-2);

\draw[->,draw=gray,  line width=.7mm] (Hidden-5) -- (Output-1);
\draw[->,draw=gray,  line width=.1mm] (Hidden-5) -- (Output-2);

\end{tikzpicture}
         \caption{Pretrained neural network before pruning.}
         \label{beforePruning}
     \end{subfigure}
     \hfill
     \begin{subfigure}[b]{0.5\textwidth}
         \centering
        \begin{tikzpicture}[scale=1.1]
\foreach \i in {1,...,\inputnum}
{
    \node[circle, 
        minimum size = 6mm,
        fill=orange!60] (Input-\i) at (0,-\i) {};
}
 
\foreach \i in {1,...,\hiddennum}
{
    \node[circle, 
        minimum size = 6mm,
        fill=teal!50,
        yshift=(\hiddennum-\inputnum)*5 mm
    ] (Hidden-\i) at (2.5,-\i) {};
}
 
\foreach \i in {1,...,\outputnum}
{
    \node[circle, 
        minimum size = 6mm,
        fill=blue!40,
        yshift=(\outputnum-\inputnum)*5 mm
    ] (Output-\i) at (5,-\i) {};
}

\draw[->,draw=gray,  line width=.9mm] (Input-1) -- (Hidden-2);
\draw[->,draw=gray,  line width=1mm] (Input-1) -- (Hidden-4);
\draw[->,draw=gray,  line width=.7mm] (Input-1) -- (Hidden-5);

\draw[->,draw=gray,  line width=.9mm] (Input-2) -- (Hidden-1);
\draw[->,draw=gray,  line width=1mm] (Input-2) -- (Hidden-3);
\draw[->,draw=gray,  line width=.5mm] (Input-2) -- (Hidden-4);

\draw[->,draw=gray,  line width=.9mm] (Input-3) -- (Hidden-5);

\draw[->,draw=gray,  line width=.5mm] (Hidden-1) -- (Output-2);

\draw[->,draw=gray,  line width=.6mm] (Hidden-2) -- (Output-1);


\draw[->,draw=gray,  line width=.9mm] (Hidden-4) -- (Output-1);
\draw[->,draw=gray,  line width=.5mm] (Hidden-4) -- (Output-2);

\draw[->,draw=gray,  line width=.7mm] (Hidden-5) -- (Output-1);

\end{tikzpicture}
         \caption{Pretrained neural network after pruning.}
         \label{fig:afterPruning}
     \end{subfigure}
    
        \caption{Visual representation of pruning a neural network. Thickness of the edges represents the magnitude of the weights. Thin edges were removed (weights set to zero) in the pruned neural network.}
        \label{fig:two graphs}
\end{figure}
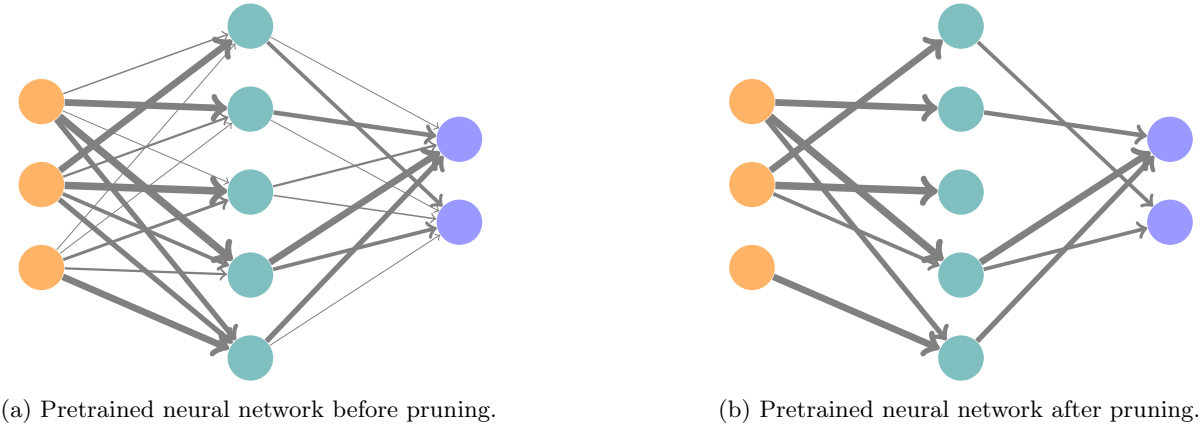

In particular, the focus of this work is on global magnitude-based pruning \cite{blalock2020state}. The magnitude of the weights is informative because it can represent the strength of the synapse between the neurons in adjacent layers within a neural network. Hence, bigger weight magnitudes can be thought to correspond to (strong) connections that are more meaningful to the overall behaviour and performance of the neural network. In a word, the magnitude of the weight may capture the importance of a connection between neurons within a neural network. Having this reasoning in mind, it is natural to consider that those weights whose absolute value is the smallest to correspond to parameters that contribute less to the final predictions of a neural network \cite{cheng2023surveydeepneuralnetwork}. A visual example  of magnitude-based pruning is presented in Figure \ref{fig:two graphs}. This motivates the essential idea of magnitude-based pruning: carrying out computations which have low impact on the performance of the model but are computationally expensive could be avoided, while simultaneously not compromising the model's performance. Of course, if these parameters are non-essential then we could also reduce memory usage if we do not need to store them.

Next, we present the setup of our magnitude-based one-shot pruning approach. Given a neural network $f:\mathcal{X}\longrightarrow[0,1]^{M\times N}$, with weights $\{w_i\}_{i=1}^{K}$ and an arbitrary sparsification hyper-parameter $\lambda\in [0,1)$, we denote by $$q_\lambda \coloneqq \texttt{quantile}( \{|w_i|\}_{i=1}^{K};\lambda)$$
the low tail $\lambda$ quantile of the absolute value of the weights in the neural network.

We denote the  sparse version of the dense neural network  by $f_{\lambda}:\mathcal{X}\longrightarrow[0,1]^{M\times N}$ , whose weights we denote by $\{w_{i,\lambda}\}_{i=1}^{K}$, where 
$$w_{i,\lambda}=\begin{cases}w_i\text{ if } |w_i|>q_\lambda\\
0 \text{ otherwise.}
\end{cases}$$

In Figure \ref{fig:two graphs} we can notice that further pruning may be possible after pruning the weights. Namely, after pruning the weights, we may prune the neurons which do not have any input or output connections by setting some additional weights to zero and we may be able to remove neurons in hidden layers from the architecture, without producing any effect on the predictions of the pruned model. In Figure \ref{fig:two graphs}, after pruning, the third neuron (from top to bottom) in the hidden layer has no connection to the neurons in the final layer. We could remove any edge that connects neurons (nodes) in the initial layer to the third neuron in the hidden layer, even if the weights have a big magnitude. Hence, the original definition of $\{w_{i,\lambda}\}_{i=1}^{K}$ may  allow to zero some additional weights leading to further sparsity.

For an $(X,Y)$ pair we denote by $\hat{Y}_{\text{full}}(X)$ the point prediction using $f$ and $\hat{Y}_{\lambda}(X)$  to the point prediction using $f_{\lambda}$. In the case where $\lambda=0$ we consider $f=f_{0}$. Of course, one of the main objectives of pruning is to leverage sparsity, so in principle one can only store the non-zero valued weights with the corresponding indices within the architecture of the neural network to reduce the memory costs and use computational tools that leverage sparsity in scientific computing settings, e.g. use sparse tensors to accelerate the prediction computations.

\subsection{The Learn then Test calibration framework}\label{section_LTT}

The Learn then Test (LTT) calibration framework  \cite{angelopoulos2022learn} is at the core of this  work. We will give a brief description of the framework adapted to this work. We have a calibration dataset $\mathcal{D}_{\text{cal}}\coloneqq\{(X_i,Y_i)\}_{i=1}^{n}$ consisting of i.i.d.  instances from a distribution $\mathbb{P}_{X,Y}$, as well as a loss function $\ell$ bounded in $[0,1]$. Our hyper-parameter space $\Lambda$ is $[0,1)$ in this case. Let $\Tilde{\Lambda}\coloneqq \{j/Q: j \in \{0,\dots, Q-1\}\}=\{\lambda_j:j\in\{0,\dots, Q-1\}\}$ (with say $Q=100$) be a discretization of $\Lambda$. We take $\lambda_j\coloneqq \frac{j}{Q}$ and consider the null hypothesis $H_{0,\lambda_j}:\mathbb{E}[\ell(Y,\hat{Y}_{\lambda_j}(X))]>\alpha$ for each $j\in \{0,\dots, Q-1\}$. Denote by $p_j$ a corresponding valid p-value for each $H_{0,\lambda_j}$. We define the empirical risk on the calibration dataset using a pruning ratio of $\lambda_j$ as $\Hat{R}_j\coloneqq \frac{1}{n}\sum_{i=1}^{n}\ell\big( Y_i, \Hat{Y}_{\lambda_j}(X_i)\big)$. In this work we will make use of different super-uniform p-values  which we explicitly mention next.

\begin{proposition}
    
 Let $\ell$ be a binary loss function (that is, taking values only in $\{0,1\}$). For any given $j\in \{0,1,\dots, Q-1\}$ we have that $p_j\coloneqq \mathbb{P}\big( Bin(n,\alpha)\leq n\Hat{R}_j\big)$ is a super-uniform p-value. That is,

$$\mathbb{P}\Big( p_j\leq u\Big| H_{0,\lambda_j}\Big)\leq u\text{ for all }u\in (0,1).$$
\end{proposition}

For a proof of this fact we refer the reader to \cite{quach2023conformal}, where the authors also work with a binary loss in a language modeling context. Relevant insights about this super-uniform p-value are also presented in \cite{bates2021distributionfree}. If the loss is more arbitrary and bounded in $[0,1]$ some options of valid p-values to use are those introduced in \cite{bates2021distributionfree,angelopoulos2022learn,alvarez2024distributionfree}.

\begin{proposition} Let $\ell$ be an arbitrary loss function bounded in $[0,1]$. Define $$\gamma(\alpha)\coloneqq\text{min}\{r\in \mathbb{N} | \quad r\geq n\alpha\}.$$ 
For any given $j\in \{0,1,\dots, Q-1\}$, 

$$p^{PRW}_{j}\coloneqq\begin{cases}
    
 \frac{\alpha(n-\lceil n\Hat{R}_j \rceil)}{n\alpha-\lceil n\Hat{R}_j \rceil}\Prob\{Bin(n,\alpha)\leq \lceil n\Hat{R}_j \rceil \} , \text{ if }\Hat{R}_j<\frac{\gamma(\alpha)-1}{n}\\\\
 
 \text{max}\{1, \frac{\alpha(n-\gamma(\alpha)+1)}{n\alpha-\gamma(\alpha)+1}\Prob\{Bin(n,\alpha)\leq \gamma(\alpha)-1\}\} , \text{ otherwise, }
 \end{cases}$$ is a super-uniform p-value. That is,

$$\mathbb{P}\Big( p^{PRW}_j\leq u\Big| H_{0,\lambda_j}\Big)\leq u\text{ for all }u\in (0,1).$$
\end{proposition}
The proof is given in \cite{alvarez2024distributionfree}.

It is also important to consider the celebrated Hoeffding-Bentkus (H-B) super-uniform p-value presented in \cite{angelopoulos2022learn, bates2021distributionfree}, which is valid for arbitrary losses bounded in $[0,1]$.

\begin{proposition}

Let $\ell$ be an arbitrary loss function bounded in $[0,1]$. For every $j\in \{0,1,\dots, Q-1\}$ define

$$p_{j}^{H-B}\coloneqq\text{min}\Big\{e\mathbb{P}\{Bin(n,\alpha)\leq \lceil n\Hat{R}_{j}\rceil \}   ,exp\{-nh_{1}(\text{min}\{\Hat{R}_{j}, \alpha \}, \alpha)\}\Big\},$$

where $h_{1}(a,b)\coloneqq alog(\frac{a}{b})+(1-a)log(\frac{1-a}{1-b})$. Then $p_{j}^{H-B}$ is a super-uniform p-value.

\end{proposition}
We will make use of these three valid p-values across the rest of this work in the experiments.

If $\Gamma\subseteq \Tilde{\Lambda}$  denotes the pruning hyper-parameters associated to the rejected null hypotheses by some family-wise error rate (FWER) controlling algorithm at level $\delta$, $\mathcal{A}=\mathcal{A}\big(p_0,\dots,p_{Q-1};\delta\big)$, then by the Theorem 1 from \cite{angelopoulos2022learn}, we have that:

$$\mathbb{P}\Big(\sup_{\lambda\in \Gamma}(\mathbb{E}[\ell(Y, \hat{Y}_{\lambda}(X))])\leq \alpha \Big)\geq 1-\delta,$$
where the supremum over the empty set is defined as $-\infty$. In the case where the loss function is monotone or approximately monotone in $\lambda$, a common election for a FWER-controlling algorithm is the fixed-sequence procedure. We present a pseudocode in the Appendix. This algorithm controls the FWER at any user specified $\delta\in (0,1)$, namely the probability of rejecting at least one true null out of the whole collection of nulls $\{H_{0,\lambda_j}\}_{j=0}^{Q-1}$ under the fixed sequence precedure is less than or equal to $\delta$. This algorithm is considered in many settings and applications \cite{Bretz2011GraphicalAF,angelopoulos2022learn, laufer2023efficiently, ringel2024early}. 
It is useful because it can leverage prior knowledge that we have about the experiment. Under many loss functions, a relationship that generally holds is that inducing higher sparsity levels usually leads to a loss degradation, thus we expect to obtain bigger risk values for bigger pruning ratios, and  this relationship is approximately monotonic.






\section{Experiments}
We will consider two settings to perform experiments in computer vision tasks: classification using the MNIST dataset \cite{lecun1998gradient} and image segmentation using the PolypGen dataset \cite{Ali_2023}.

\subsection{Classification}
\begin{figure}[H]
    \centering
\begin{tikzpicture}[x=1.9cm, y=1.1cm, >=stealth]

\newcommand{\inputlayer}{4}
\newcommand{\hiddenlayersize}{5}
\newcommand{\outputlayer}{3}  

\colorlet{inputcolor}{blue!30}
\colorlet{hiddencolor}{green!30}
\colorlet{outputcolor}{red!30}

\foreach \i in {1,...,\inputlayer}
  \node[circle, draw=black, fill=inputcolor, minimum size=0.9cm] (I-\i) at (0,-\i*1.5) {$x_\i$};
   
\node at (0,-\inputlayer*1.1-0.75) {$\vdots$};

\node[circle, draw=black, fill=inputcolor, minimum size=0.8cm] (I-4) at (0,-4*1.5) {$x_{784}$};

\foreach \layer [count=\l] in {1,2}
{
  \foreach \n in {1,...,\hiddenlayersize}
  {
    \pgfmathtruncatemacro{\yshift}{(\inputlayer*1.5-1)/2 - (\hiddenlayersize-1)/2}
    \node[circle, draw=black, fill=hiddencolor, minimum size=0.99cm] (H\l-\n) at (\l*2,-\n*1.5+\yshift) {$h_{\n}^\l$};
    \node at (\layer*2,-6.65) {$\vdots$};
  }
}

\node[circle, draw=black, fill=hiddencolor, minimum size=0.8cm] (H1-5)  at (1*2,-5*1.5) {$h_{128}^{1}$};

\node[circle, draw=black, fill=hiddencolor, minimum size=0.8cm] (H2-5) at (2*2,-5*1.5) {$h_{128}^{2}$};

\node[circle, draw=black, fill=outputcolor, minimum size=0.8cm] (O-1) at (2*2+2,-1.5) {$y_0$};
\node[circle, draw=black, fill=outputcolor, minimum size=0.8cm] (O-2) at (2*2+2,-3.0) {$y_1$};

\node at (2*2+2,-4.1) {$\vdots$};
 
\node[circle, draw=black, fill=outputcolor, minimum size=0.8cm] (O-3) at (2*2+2,-5.5) {$y_9$};  

\foreach \i in {1,...,\inputlayer}
  \foreach \j in {1,...,\hiddenlayersize}
    \draw[->] (I-\i) -- (H1-\j);

\foreach \j in {1,...,\hiddenlayersize}
  \foreach \k in {1,...,\hiddenlayersize}
    \draw[->] (H1-\j) -- (H2-\k);

\foreach \j in {1,...,\hiddenlayersize}
  \foreach \i in {1,2,3}  
    \draw[->] (H2-\j) -- (O-\i);

\node[align=center, above] at (0,-1) {Input\\Layer};
\node[align=center, above] at (2,-1) {Hidden\\Layer 1};
\node[align=center, above] at (2*1+2,-1) {Hidden\\Layer 2};
\node[align=center, above] at (2*2+2,-1) {Output\\Layer};
\end{tikzpicture}
\caption{Neural network architecture for our MNIST experiments.}
\label{network_example}
\end{figure}
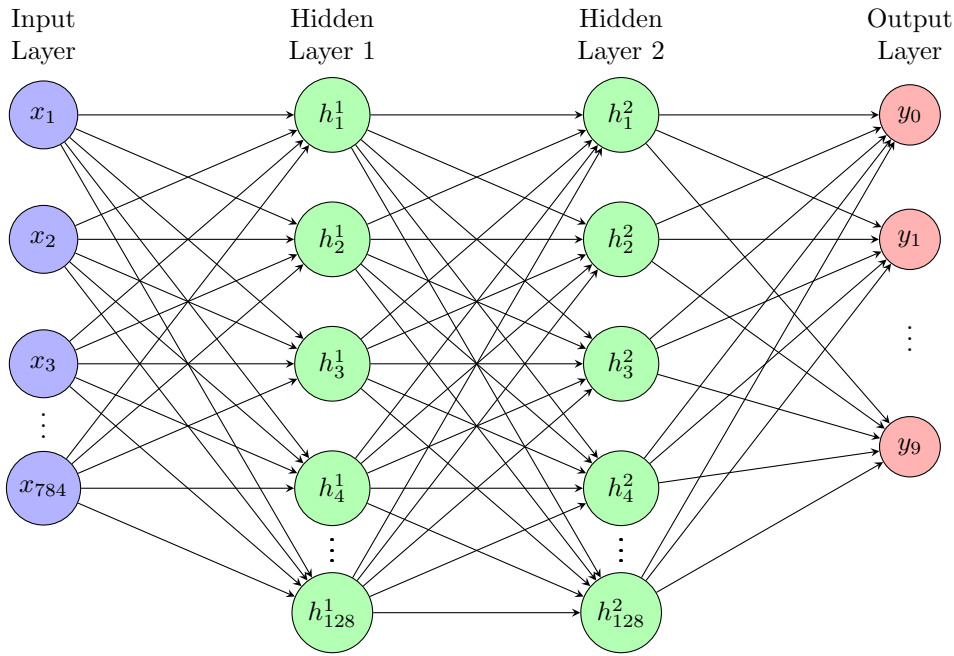

\begin{figure}[H]
\begin{subfigure}{.5\textwidth}
  \centering
\includegraphics[width=.95\linewidth]{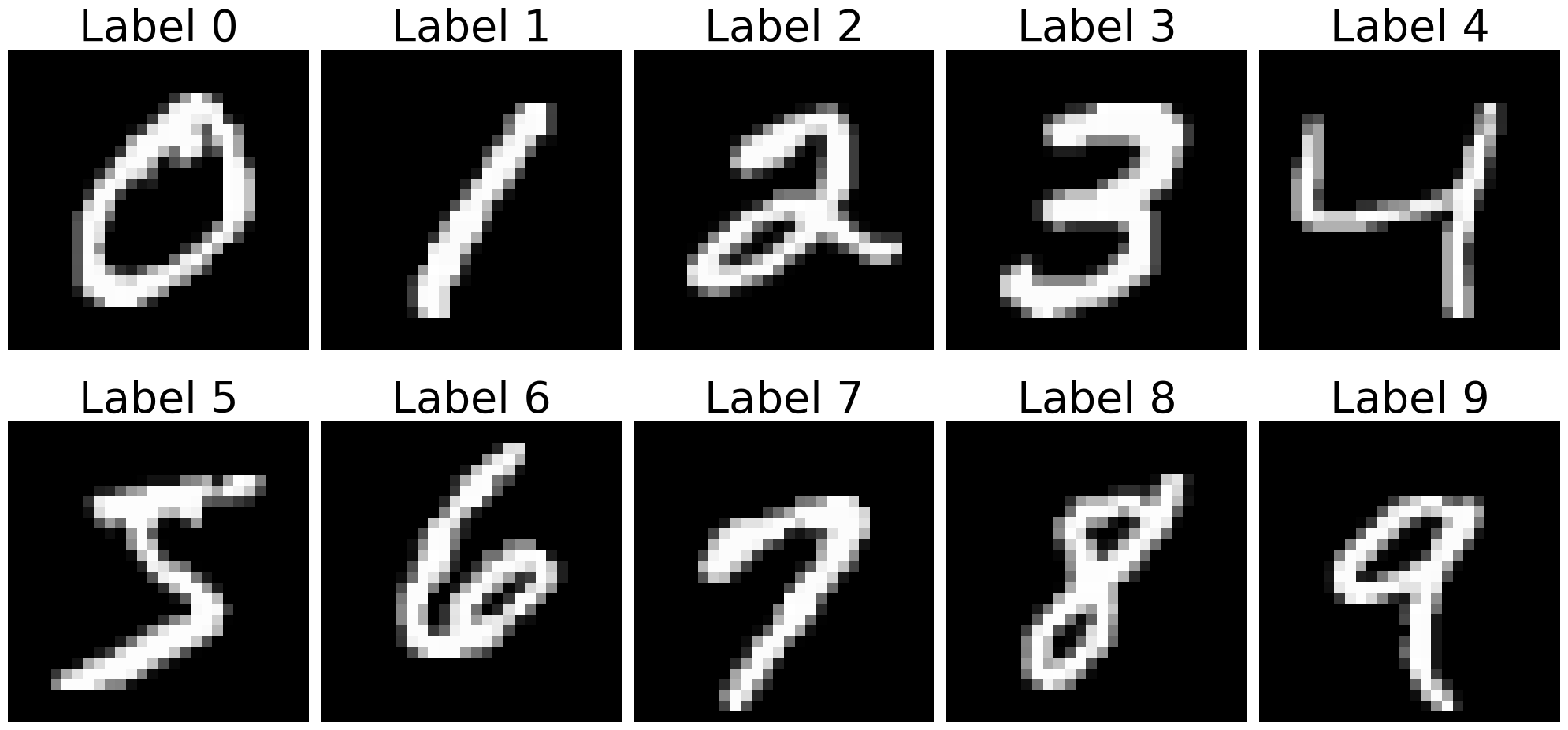}
  \caption{Sample of instances of different classes in the MNIST dataset.}
  \label{mnist}
\end{subfigure}%
\quad
\begin{subfigure}{.35\textwidth}
\centering

\includegraphics[width=.95\textwidth]{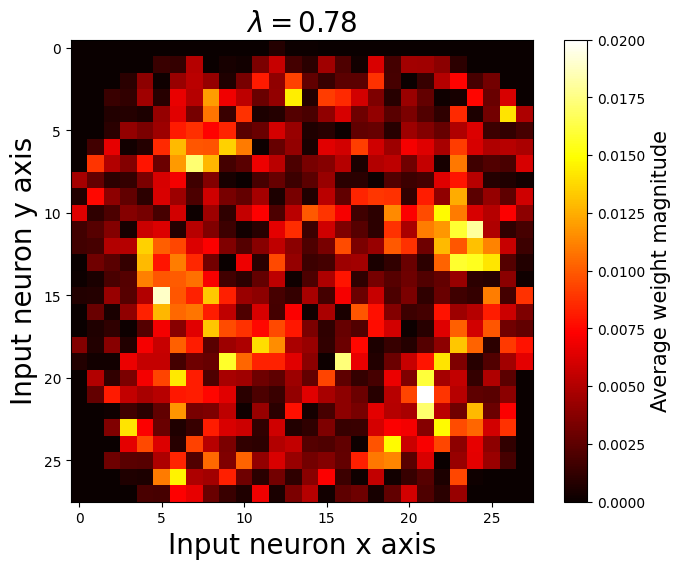}
  \caption{Average magnitude of the weights  connecting the input layer with the first hidden layer with a pruning ratio of $\lambda=0.78$.}
  \label{prunedWeights}
\end{subfigure}
\caption{Pixel representations of the inputs in the dataset \ref{mnist} and average magnitude weight map in a pixel representation \ref{prunedWeights}.}
\label{mnist_dataset_figures}
\end{figure}

We will consider the MNIST dataset \cite{lecun1998gradient} which forms part of Keras, consisting of 70,000 labeled $28\times28$ pixel images containing digits in $\{0,1,\dots, 9\}$. For a given pair $(X,Y)$, $X$ denotes a $28\times 28$ pixel image and $Y$ the digit in that image. See Figure \ref{mnist}. We split the data into train, calibration and validation datasets, containing $60,000$ images for training, $9,000$ for calibrating and $1,000$ images for testing. We use a neural network consisting of an input layer with 784 neurons (corresponding to each pixel of the image), 2 hidden layers with 128 neurons each and an output layer with 10 neurons (corresponding to each of the digits). This neural network is represented in Figure \ref{network_example}. We use ReLU activation functions to connect the first three layers and a soft-max activation function to restrict the output of the neurons in the last layer to the interval $(0,1)$. Taking into account weights and biases, the neural network has a total of 118,282 parameters.

Pixels in the corners of the images within the MNIST dataset are usually not used to write the numbers, this is something that magnitude-based pruning can capture. We see this reflected in Figure \ref{prunedWeights}, where on average, the weights in the corners have the smallest magnitude. Thus, visual attention in the corners of the image is non-informative to the predictions of the neural network. For further insights about how magnitude-based pruning can provide informative features about visual attention in a neural network we refer the reader to \cite{han2015learning}.

Next, we present some experimental examples considering different loss functions and different formulations within the MNIST context using our confident magnitude-based pruning framework.

\subsubsection{Pruning with labeled data}

\begin{figure}[H]
    \centering
    \includegraphics[width=.5\linewidth]{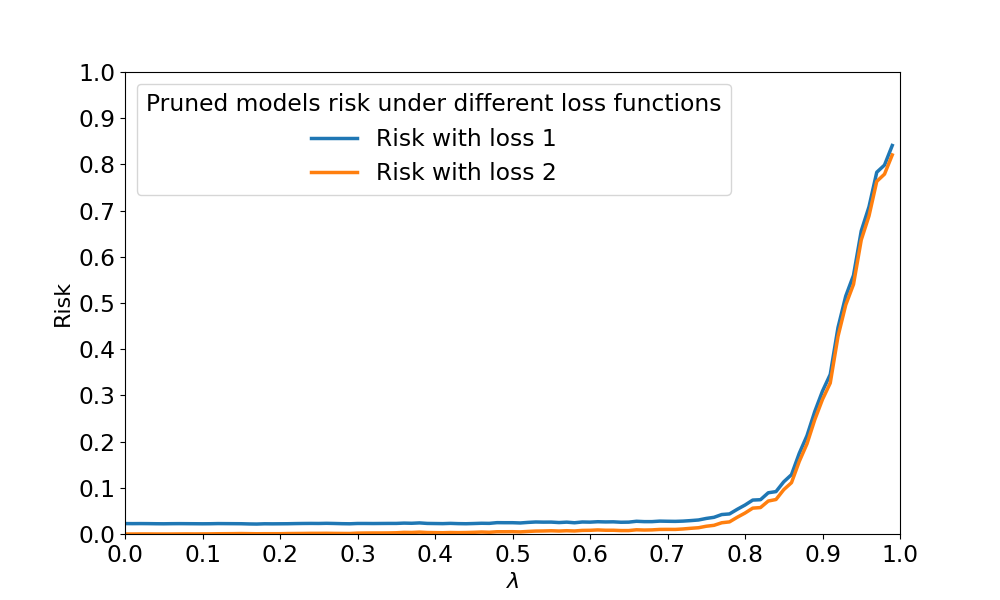}
    \caption{Risk as a function of the pruning ratio in the calibration dataset.}
    \label{figure_monotonicity}
\end{figure}

 As a first example, we will consider the following loss function for a given $(X,Y)\sim \mathbb{P}_{X,Y}$:

 \begin{equation}
    \ell\big(Y, \Hat{Y}_{\lambda}(X)\big)\coloneqq \mathbbm{1}\{Y\neq \Hat{Y}_{\lambda}(X)\}\label{loss_1},
 \end{equation}

where $\mathbbm{1}\{\cdot\}$ denotes the indicator function. The risk that we control is  $$\mathbb{E}[\ell\big(Y, \Hat{Y}_{\lambda}(X)\big)]=\mathbb{P}\big( Y\neq \Hat{Y}_{\lambda}(X)\big)=1-\mathbb{P}\big( Y= \Hat{Y}_{\lambda}(X)\big),$$ which is one minus the accuracy. We can appreciate the approximate monontonicity of this risk in the calibration dataset in Figure \ref{figure_monotonicity} in the blue line.

Another alternative formulation is to consider a loss function given by
\begin{equation}
   \Big( \mathbbm{1}\{Y\neq \Hat{Y}_{\lambda}(X)\}-\mathbbm{1}\{Y\neq \Hat{Y}_{\text{full}}(X)\}  \Big)_{+}\label{loss_2},
\end{equation}

with $(x)_{+}\coloneqq \text{max} (x,0) \text{ for every }x\in \mathbb{R}$, which is a more relaxed loss because it does not penalize in the case where the pruned model makes correct predictions and the full model classifies the digit incorrectly. Note that the loss \ref{loss_2} is also binary. The risk under loss \ref{loss_2} is given by $\mathbb{P}\Big(\{Y\neq \Hat{Y}_{\lambda}(X)\}\cap \{Y= \Hat{Y}_{\text{full}}(X)\}\Big)$.

When comparing losses \ref{loss_1} and \ref{loss_2} in terms of their expected values, by the monotonicity of the probability measure we have that 
$$ \mathbb{P}\Big(\{Y\neq \Hat{Y}_{\lambda}(X)\}\cap \{Y= \Hat{Y}_{\text{full}}(X)\}\Big)\leq \mathbb{P}\Big(Y\neq \Hat{Y}_{\lambda}(X)\Big).$$

 A similar setting in an NLP application can be found in \cite{laufer2023efficiently}. Therefore, for any given performance degradation tolerance threshold $\alpha$ and any $\delta\in (0,1)$ we would expect the fixed sequence procedure to produce higher pruning ratios under the loss \ref{loss_2} than those obtained for the loss \ref{loss_1}. This observation can be verified in the Table \ref{calibration_results} as well as in an empirical level using the calibration dataset in Figure \ref{figure_monotonicity}. Both risks are approximately non-decreasing in $\lambda$.

\begin{table}[H]
  \centering
  \caption{Calibration results with labeled data using the fixed sequence procedure  with different loss functions.}
  \begin{subtable}[t]{0.45\linewidth}
    \centering
    \caption{Using loss \ref{loss_1}.}

\begin{tabular}{ |c|c|c| }
\hline
$\alpha$ & $\delta$ & $\lambda$ \\
\hline
\multirow{3}{*}{\centering0.03} & \multirow{1}{*}{0.05} & 0.65 \\
& 0.10 &0.68\\
& 0.20 & 0.72\\
\cline{2-3} 
\hline
\multirow{3}{*}{\centering0.04} & \multirow{1}{*}{0.05} & 0.76 \\
& 0.10 &0.76\\
& 0.20 & 0.76\\
\cline{2-3} 
\hline
\multirow{3}{*}{\centering0.05} & \multirow{1}{*}{0.05} & 0.78 \\
& 0.10 &0.78\\
& 0.20 & 0.78\\
\cline{2-3} 
\hline
\end{tabular}
  \end{subtable}
  \begin{subtable}[t]{0.45\linewidth}
    \centering
    \caption{Using loss \ref{loss_2}.}

\begin{tabular}{ |c|c|c| }
\hline
$\alpha$ & $\delta$ & $\lambda$ \\
\hline
\multirow{3}{*}{\centering0.03} & \multirow{1}{*}{0.05} & 0.78 \\
& 0.10 &0.78\\
& 0.20 & 0.78\\
\cline{2-3} 
\hline
\multirow{3}{*}{\centering0.04} & \multirow{1}{*}{0.05} & 0.79 \\
& 0.10 &0.79\\
& 0.20 & 0.79\\
\cline{2-3} 
\hline
\multirow{3}{*}{\centering0.05} & \multirow{1}{*}{0.05} & 0.80 \\
& 0.10 &0.80\\
& 0.20 & 0.80\\
\cline{2-3} 
\hline
\end{tabular}
  \end{subtable}
  \label{calibration_results}
\end{table}

Overall, the obtained sparsification hyper-parameters that we obtained in the calibration procedure showed more sensitivity to $\alpha$ than to $\delta$. In many cases, the achieved sparsity was the same for different values of $\delta$ when fixing the performance degradation tolerance.

\subsubsection{Pruning with unlabeled data}

When considering pruning neural networks, a typical perspective is to compare the pruned model's predictions with the predictions of the dense network. This motivates us to consider a setting of pruning with unlabeled data, because comparing the predictions of the pruned model with the full model does not require to know the true label or the true outcome to be predicted. This approach is also motivated by situations in with the user already has a neural network that satisfies the necessary requirements for making predictions, but one would like to make the network cheaper to store and more efficient to make predictions. Under this risk control formulation we will consider pruning neural networks while having statistical guarantees of making the same predictions.

This approach was motivated by the recent success of  efficient LLMs with rigorous statistical guarantees and unlabeled text generation in \cite{schuster2022confident}, where the authors also consider calibrating with unlabeled datasets. For our context we will consider the loss

\begin{equation}
    \ell\big( \Hat{Y}_{\text{full}}(X), \Hat{Y}_{\lambda}(X)\big)\coloneqq \mathbbm{1}\{\Hat{Y}_{\text{full}}(X)\neq \Hat{Y}_{\lambda}(X)\},
    \label{unlabeled_loss}
\end{equation}

 so that we control the probability of making different predictions with the pruned model, namely our risk is:

$$\mathbb{P}\big(\Hat{Y}_{\text{full}}(X)\neq \Hat{Y}_{\lambda}(X)\big).$$

The motivation behind this approach is to prune a model until the predictions of the pruned model differ from the predictions of the dense model with some user specified probability. Note that we do not need to consider the true label corresponding to a given image $X$.

The most appropriate way perform checks on the coverage of the calibration is to perform several iterations of the splitting repeating the calibration procedure \cite{anastasiosconfromal}, though this can be computationally expensive.  Hence, we consider doing a sanity check of the calibration through the bootstrap in a single iteration of the calibration step as presented in Algorithm \ref{algorithm2} in the Appendix section.

\begin{figure}[H]
\centering
\begin{subfigure}{.5\textwidth}
  \centering
  \includegraphics[width=.95\linewidth]{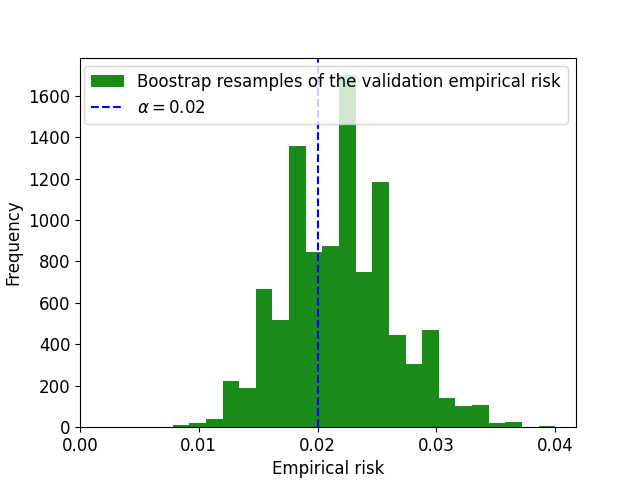}
  \caption{Empirical risks with $\lambda=0.73$.}
  \label{fig:sub1}
\end{subfigure}%
\begin{subfigure}{.5\textwidth}
  \centering
  \includegraphics[width=.95\linewidth]{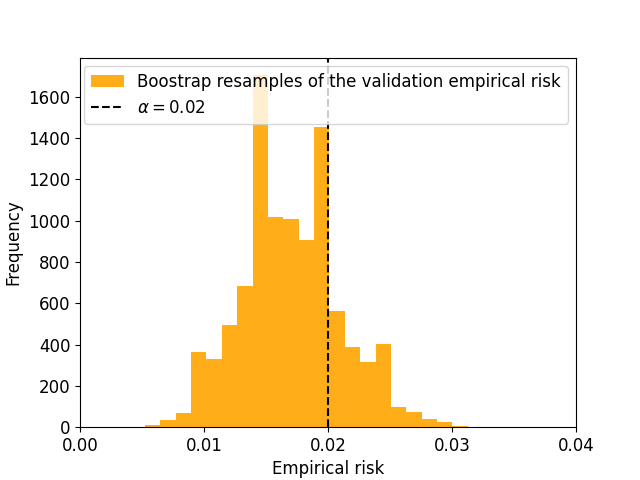}
  \caption{Empirical risks with $\lambda=0.71$.}
  \label{fig:sub2}
\end{subfigure}
\caption{Bootstrap distribution of the empirical risk with the loss function \ref{unlabeled_loss} on the validation dataset calibrating the pruning ratio with $\alpha=0.02$ and $B=10,000$ bootstrap resamples. Figure \ref{fig:sub1} corresponds to pruning with a naïve approach without accounting for uncertainty. Figure \ref{fig:sub2} was obtained calibrating with  fixed sequence testing taking $\delta=0.10$.}
\label{bootstrap_dist}
\end{figure}
We can appreciate the resulting differences between a naïve calibration approach which terminates the pruning after the first time that the empirical risk gets below a desired threshold and calibrating with via the LTT calibration framework in Figure \ref{bootstrap_dist}. Under the naïve approach \ref{fig:sub1} we get a bigger pruning ratio, but the distribution of the empirical risk accumulates more probability mass above $\alpha$  compared to \ref{fig:sub2} which takes into account uncertainty rigorously. Moreover, taking an overly conservative approach to terminate the pruning could result to under-pruned models.

\subsubsection{Pruning with selective predictions}

Selectivity is a helpful approach to make sure that the model makes predictions when being sufficiently confident about them, and abstaining to predict otherwise. In a pruning context, selective predictions might be of interest to determine how confident does a pruned model remain after inducing sparsity. An approach to take into account the confidence of the model (measured through the softmax output of the predicted probabilities of our neural network) is through calibrating a selective prediction hyper-parameter. 

In fact, we have two hyper-parameters to calibrate. Indeed, the LTT framework is  useful because it can work with more than one hyper-parameter. In this particular calibration setup, we consider $\lambda_1,\lambda_2\in [0,1)$, where $\lambda_1$ corresponds to a selective threshold and $\lambda_2$ corresponds to the pruning ratio. Hence, the hyper-paramter space now becomes $\Lambda\coloneqq [0,1]\times[0,1]$. We consider making predictions with

$$\Hat{Y}_{\lambda_1,\lambda_2}(X)\coloneqq \begin{cases}
    \hat{Y}_{\lambda_2}(X)& \text{if } \hat{Y}_{\lambda_2}(X)>\lambda_1 \\

    \emptyset,& \text{otherwise.}
\end{cases}$$

On the one hand, we prefer to predict as often as possible, because a model that makes more predictions will provide more utility to make decisions in practice. Of course, making more predictions has a tradeoff with accuracy, because making more predictions forces the model to make predictions even in hard instances. On the other hand, abstaining may allow the model to make predictions when it is more confident. Thus, we prefer smaller values of $\lambda_1$ (leading to making predictions more often) as long as that does not compromise the accuracy below a user-specified tolerance.

We will control the probability of wrong predictions conditional on predicting, namely, our risk is given by:

$$\mathbb{P}\Big(\Hat{Y}_{\lambda_1,\lambda_2}(X)\neq Y \big|  \Hat{Y}_{\lambda_1,\lambda_2}(X)\neq \emptyset \Big).$$
Recall that our calibration dataset is denoted by $\mathcal{D}_{\text{cal}}\coloneqq \{(X_i,Y_i)\}_{i=1}^{n}$. Let $$\mathcal{I}_{\lambda_1, \lambda_2}\coloneqq \{ i\in \{1,\dots, n\}:\Hat{Y}_{\lambda_1,\lambda_2}(X_i) \neq \emptyset \}$$  be the set of indices of the calibration dataset where the model does not abstain, and let $n_{\lambda_1,\lambda_2}\coloneqq \big|\mathcal{I}_{\lambda_1, \lambda_2} \big|\in\{0,\dots, n\}$ denote the number of elements in $\mathcal{I}_{\lambda_1, \lambda_2}$.
Our empirical risk given a selectivity threshold $\lambda_1\in(0,1)$ and a pruning ratio $\lambda_2\in [0,1)$ is  defined as 

$$\Hat{R}(\lambda_1,\lambda_2)\coloneqq \frac{1}{n_{\lambda_1,\lambda_2}}\sum_{i\in \mathcal{I}_{\lambda_1, \lambda_2}} \mathbbm{1}\big\{ \hat{Y}_{\lambda_2}(X_i)\neq Y_{i}\big\},$$
where we can clip the pruning ratio in the calibration or explore for small selective hyper-parameters to avoid degeneracy. We can also monitor the fraction of abstentions which is given by the fraction $\frac{n-n_{\lambda_1,\lambda_2}}{n}$.

We consider the fallback procedure in a graphical approach \cite{wiens2005fallback, Bretz2011GraphicalAF, angelopoulos2022learn} as described in the Appendix of \cite{angelopoulos2022learn} to calibrate with two hyper-parameters. Concretely, we take  $\Tilde{\Lambda}_1\coloneqq \{ \lambda_{1,k}| k\in \{1,\dots, J\}\}\subseteq (0,1)$ as our selectivity parameters to search with $\lambda_{1,1}<\lambda_{1,2}<\dots<\lambda_{1,J}$, and $\Tilde{\Lambda}_2=\{\lambda_{2,j}:j\in \{0,1,\dots, T\}\}$ with $\lambda_{2,j}\coloneqq \frac{j}{Q}\text{ for each } j\in \{0,\dots, T\}$, so that $\Tilde{\Lambda}\coloneqq \Tilde{\Lambda}_1 \times \Tilde{\Lambda}_2$ forms our search space, for some $J,T,Q\in\mathbb{N}$, such that $T<Q$ (e.g. $T=80<Q=100$ to avoid degeneracy). Given a threshold $\alpha\in (0,1)$, we consider null hypotheses of the form $H_{\lambda_{1,k}, \lambda_{2,j}}:\mathbb{P}\Big(\Hat{Y}_{\lambda_{1,k}, \lambda_{2,j}}(X)\neq Y \big|  \Hat{Y}_{\lambda_{1,k}, \lambda_{2,j}}(X)\neq \emptyset \Big)>\alpha$ for every element in $\Tilde{\Lambda}$.

The basic idea in the graphical approach for the fallback procedure in this context is to think about the null hypotheses as nodes within a graph. We allocate an initial budget of $\delta/J$ on each of the nulls of the form $H_{\lambda_{1,k},\lambda_{2,0}}$ for each $k\in\{1,\dots, J\}$ and we assign a weight of 1 to the edges of the graph connecting $H_{\lambda_{1,k},\lambda_{2,j}}$ to $H_{\lambda_{1,k},\lambda_{2,j+1}}$ for every $k\in\{1,\dots, J\} \text{ and every } j\in\{0,1,\dots, T-1\}$ to redistribute the allocated budget. The initial budget allocations for each node $\big\{\delta_{k,j}|k\in \{1,\dots, J \}\text{ and }j\in \{0,\dots,  T\}\big\}$ are given by $\delta_{k,0}=\delta/J$ for every $k\in \{1,\dots, J \}$ and equal zero for the rest of the null hypotheses.  We include a pseudocode of the fallback procedure to make the calibration in this context in the Appendix of this work. The algorithm controls the FWER at level $\delta$.

\begin{figure}[H]
\begin{subfigure}{.5\textwidth}
  \centering
\includegraphics[width=.95\linewidth]{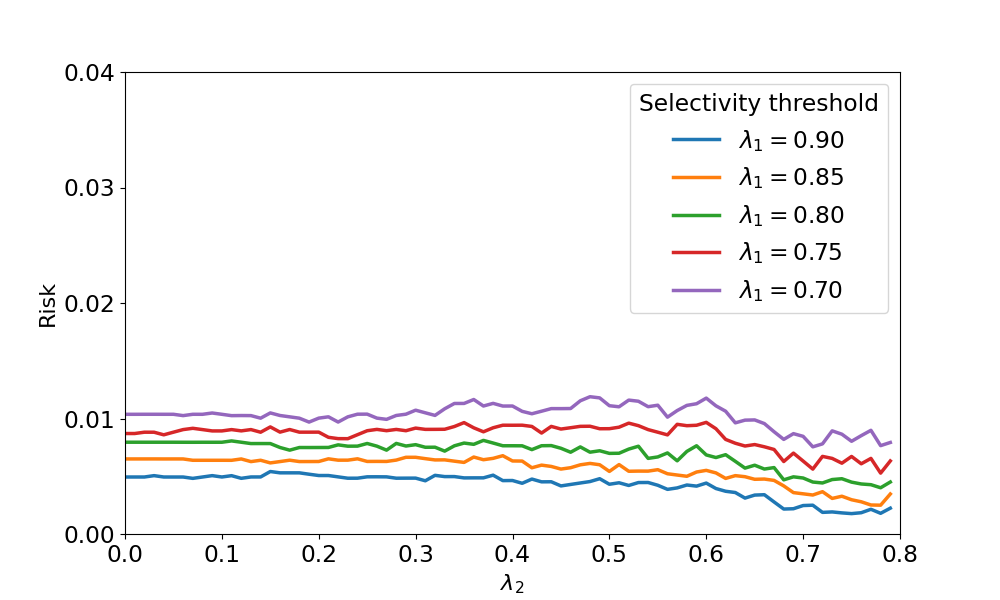}
  \caption{Risk for different hyper-parameters.}
  \label{selective_pruning}
\end{subfigure}%
\quad
\begin{subfigure}{.5\textwidth}
\centering
\includegraphics[width=.94\textwidth]{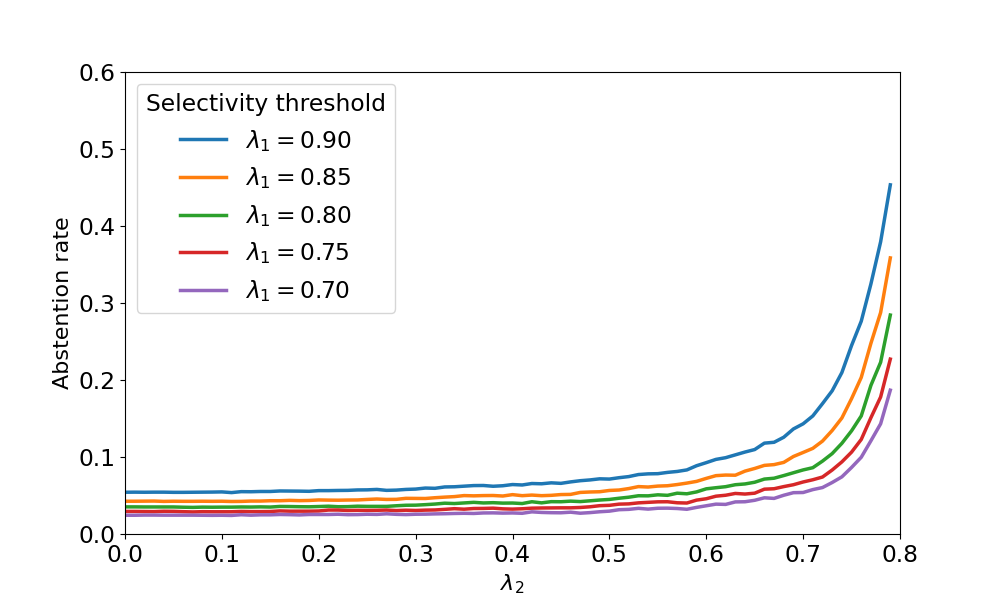}
  \caption{Abstention rates.}
  \label{abstention_rates}
\end{subfigure}
\caption{Risk \ref{selective_pruning}  and abstention rates \ref{abstention_rates}  in the calibration dataset.}
\label{Selective_classification}
\end{figure}

In Figure \ref{selective_pruning} we can observe the effect of making selective predictions compared to not abstaining at all as we have seen in previous experiments: the risk gets substantially smaller when making selective predictions. Moreover, by increasing abstention rates for higher sparsification hyper-parameters  \ref{abstention_rates}, the risk remains relatively stable across the sparsification hyper-parameter. As we would expect, higher selectivity thresholds lead to higher abstention rates and smaller risk levels.

\subsection{Pruning image segmentation neural networks}

For this experiment we train a unet architecture \cite{ronneberger2015unet} consisting of $13,846,273$ parameteters. We use the curated polyp segmentation dataset from \cite{Ali_2023}. Given an $(X,Y)$ pair, $X$ denotes a colonoscopy image in the RGB system, $X\in \mathcal{X}\subseteq\mathbb{R}^{M \times N\times 3}$, and $Y\subseteq\{1,\dots, M\}\times\{1,\dots, N\}$ denotes the pixel coordinates of the image where there is a tumour. We use a neural network $f:\mathcal{X}\longrightarrow[0,1]^{M \times N}$, and we denote by $f(X)$ the output of the neural network. In this case, $f(X)_{i,j}$ denotes the estimated probability that there is a tumour in the pixel $(i,j)$ of the image. Then our prediction set of the pixels where there is a tumour is given by

$$\Hat{Y}_{\text{full}}(X)\coloneqq\{(i,j)\in \{1,\dots, M\}\times\{1,\dots, M\}:f(X)_{i,j}\geq 1-\beta\},$$
where  we take $\beta\in (0,1)$ as a given threshold.  Analogously, we consider prediction sets of the form 
$$\hat{Y}_{\lambda}(X)\coloneqq\{(i,j)\in \{1,\dots, M\}\times\{1,\dots, M\}:f_{\lambda}(X)_{i,j}\geq 1-\beta\},$$
to make predictions with the pruned model. 
We compare the the prediction masks of the full model to the predictions of the pruned unet models through the Intersection-over-Union (IoU) of the predictions, namely,
$$IoU\big(\hat{Y}_{\text{full}}(X), \hat{Y}_{\lambda}(X)\big)\coloneqq \frac{\big|\hat{Y}_{\lambda}(X)\cap\hat{Y}_{\text{full}}(X)\big|}{\big|\hat{Y}_{\lambda}(X)\cup\hat{Y}_{\text{full}}(X)\big|},$$
where we define the function $| \cdot |$ as the number of elements in a set. The IoU takes values in $[0,1]$. Our convention is to set the $IoU$ equal to 1 in the degenerate case where the denominator is equal to zero. We consider the loss function

$$\ell\big(\hat{Y}_{\text{full}}(X), \hat{Y}_{\lambda}(X)\big)\coloneqq 1-IoU\big(\hat{Y}_{\text{full}}(X), \hat{Y}_{\lambda}(X)\big).$$

\begin{figure}[H]
  \begin{minipage}[b]{.5\linewidth}
    \centering
    \includegraphics[width=0.9\linewidth]{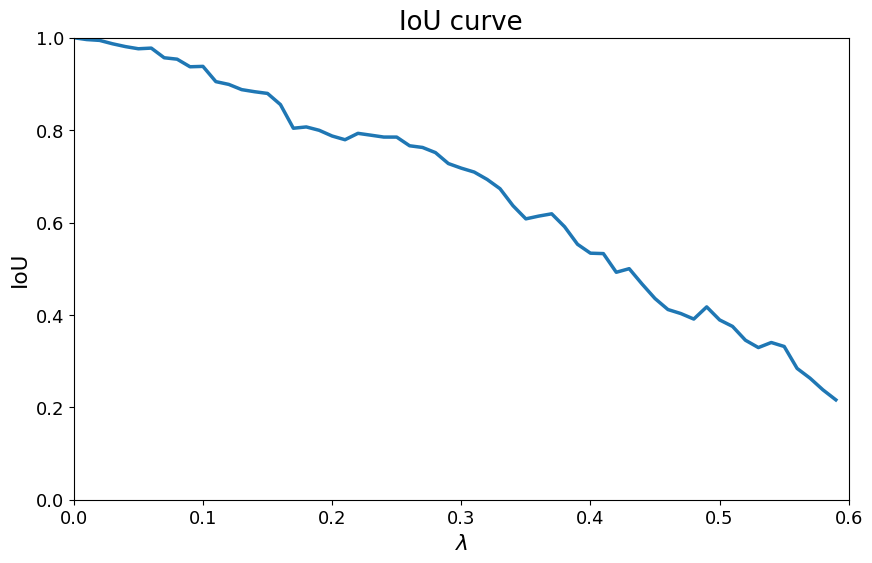}
    \captionof{figure}{IoU in the calibration dataset}
    \label{iou_calibration}
  \end{minipage}\hfill
  \begin{minipage}[b]{.45\linewidth}
    \centering
   \begin{tabular}{ |c|c|c| }
\hline
$\alpha$ & $\delta$ & $\lambda$ \\
\hline
\multirow{3}{*}{\centering0.02} & \multirow{1}{*}{0.05} & 0.02\\
& 0.10 &0.02\\
& 0.20 & 0.02\\
\cline{2-3} 
\hline
\multirow{3}{*}{\centering0.05} & \multirow{1}{*}{0.05} & 0.06 \\
& 0.10 &0.06\\
& 0.20 & 0.06\\
\cline{2-3} 
\hline
\multirow{3}{*}{\centering0.10} & \multirow{1}{*}{0.05} & 0.10 \\
& 0.10 &0.10\\
& 0.20 & 0.10\\
\cline{2-3} 
\hline
\end{tabular}
    \captionof{table}{Calibration results for the image segmentation task using the fixed sequence procedure. The calibration produced the same sparsification hyper-parameters when using the H-B valid p-value and the PRW valid p-value.}
    \label{calibration_segmentation}
  \end{minipage}
\end{figure}

This loss takes values in $[0,1]$. Thus, we need to consider an adequate valid p-value as discussed in the Section \ref{section_LTT}. The intuition is that higher pruning ratios reduce the value of the IoU, because the underlying neural networks differ more. Hence, we consider the fixed sequence testing with the H-B super-uniform p-value and we also consider using the PRW super-uniform p-value.  Implementing the calibration procedure requires to change line 4 of the Algorithm \ref{algorithm} from the Appendix using one of these valid p-values. Results on the calibration procedure are shown in Table \ref{calibration_segmentation}. In Figure \ref{iou_calibration} we can observe an approximately non-increasing relationship between the IoU and the pruning ratio using the calibration dataset consisting of 465 images.

Indeed, global-magnitude based pruning did not achieve great compression rates in this experiment, as the results in the Table \ref{calibration_segmentation} demonstrate. For small values of the pruning ratio the pruned model's predictions begin to deviate from the dense models predictions rapidly as we gradually increase the sparsity.  Another approach which could potentially achieve higher pruning ratios would be to calibrate $\beta$ in the pruned model rather than taking it fixed as given by the dense model.

\begin{figure}[H]
    \centering
\includegraphics[width=.5\linewidth]{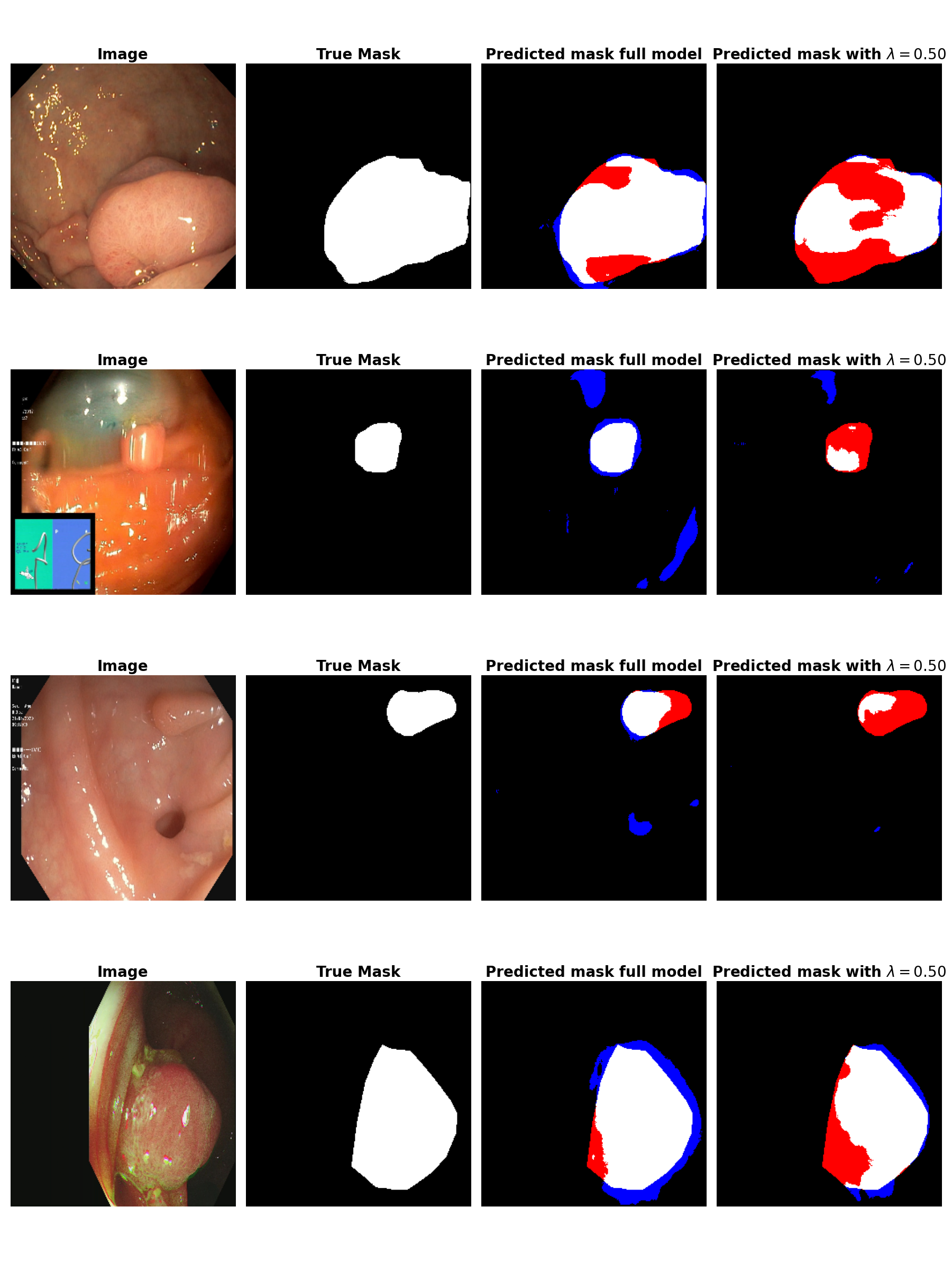}
    \caption{Sample of instances 
 with a pruning ratio $\lambda=0.50$. Black pixels are true negatives, white pixels are true positives, red pixels are false negatives and blue pixels are false positives.}
    \label{figure_polyps}
\end{figure}

In Figure \ref{figure_polyps} we can see that global magnitude based pruning does not allow to induce big sparsification for small risk control thresholds, as compared to the experiment results using the MNIST dataset for classification, where compression ratios bigger than $50\%$ were achieved with negligible performance degradation levels. For big sparsity ratios, the predicted masks of the dense and the pruned neural networks differ widely.

This experiment allows us to discern a very important matter in pruning. The fact that we are not able to compress a model to a big pruning ratio does not necessarily mean that the model is not over-parametrized. It is crucial to consider if the pruning approach is the appropriate one to achieve a big compression without compromising performance. We should highlight that the experiments were carried out in a global magnitude-based fashion but maybe using a different pruning strategy, especially one that leverages the structure of the unet architecture could potentially lead to higher compression rates. Moreover, when considering model compression by inducing sparsity through pruning to an alternative approach of using smaller but dense neural networks, the comparison should account for the pruning approach \cite{zhu2017prune}.

This experiment could be extended with different formulations. For example, with the false negative rate (FNR) as our loss function, or using labeled data, or calibrating $\beta$ rather than fixing it. The experiment did not need to have the corresponding true masks for the images. Further experiments could consider pruning and using a loss function that penalizes based on certain criterion that measures how much does the pruned model's predicted masks deviate from the true mask.

\section{Conclusions}
This work presented a practical approach to carry out magnitude-based pruning with uncertainty-aware stopping rules in computer vision. The approach is useful especially for those one-shot pruning methods that begin with a dense neural network and gradually induce sparsity in the model. We carried out experiments in different computer vision settings, illustrating that the approach can be implemented in practice with great versatility. Furthermore, we observed that the calibrated sparsification hyper-parameter presented greater sensitivity to the performance degradation tolerance than to the error budget in the marginal risk control guarantees.

\section*{Acknowledgements}
I want to thank Professor Miguel Angel Mota  from ITAM as well as Anastasios Nikolas Angelopoulos for inspiring me to do research. Without their encouragement, this work would not have been possible. I also thank Cristina Diaz Faloh for insightful conversations about the fascinating idea of pruning neural networks at the theoretical neuroscience workshop  in the Abdus Salam ICTP.

\printbibliography

\section*{Appendix}

We provide a list of algorithms that we use to carry out our experiments.

\begin{algorithm}[H]
    \caption{Fixed Sequence Algorithm adapted to our framework}\label{algorithm}
    \hspace*{\algorithmicindent} \textbf{Inputs:} $\mathcal{D}_{\text{cal}}\coloneqq\{(X_i,Y_i)\}_{i=1}^{n}$ our calibration dataset, $\Tilde{\Lambda}=\{\lambda_1, \dots, \lambda_{Q-1}\}$, $\delta\in (0,1)$  a level to control the FWER, $\alpha\in (0,1)$ our desired risk control threshold. \\
    \hspace*{\algorithmicindent} 
    \textbf{Output:} $\Gamma$ pruning ratios whose associated null hypothesis was rejected.
    \begin{algorithmic}[1]
    \State $\Gamma\gets  \varnothing$\textcolor{olive}{ \Comment{initialize the pruning parameters}}
\For{$j\text{ in } 0,\dots ,Q-1$}
\State $\Hat{R}_j\gets \frac{1}{n}\sum_{i=1}^{n}\ell\big( Y_i, \Hat{Y}_{\lambda_j}(X_i)\big)$
\State $p_{j}\gets \mathbb{P}\big( Bin(n,\alpha)\leq n\Hat{R}_j\big)$\textcolor{olive}{ \Comment{Compute a super-uniform p-value associated to $H_{0,\lambda_j}$.}}
\If{$p_{j}\leq \delta$}
\State $\Gamma\gets \Gamma\cup \{\lambda_j\}$\textcolor{olive}{ \Comment{$\Gamma$ gets updated since $H_{0,\lambda_j}$ is rejected}}
\Else
\State \textbf{break for}
\EndIf
\EndFor
\\
\Return $\Gamma$
\end{algorithmic}
\end{algorithm}

\begin{algorithm}[H]
    \caption{Sampling from the bootstrap distribution of the empirical risk on the validation dataset to check the calibration \cite{hastie2009elements}}\label{algorithm2}
    \hspace*{\algorithmicindent} \textbf{Inputs:} $\mathcal{D}_{\text{cal}}\coloneqq\{(X_i,Y_i)\}_{i=1}^{n}$, $\mathcal{D}_{\text{val}}\coloneqq\{(X_i,Y_i)\}_{i=n+1}^{m+n}$,$\Tilde{\Lambda}$, $\delta\in (0,1)$  a level to control the FWER, $\alpha\in (0,1)$ our desired risk control threshold, $B$ a number of bootstrap resamples \\
    \hspace*{\algorithmicindent} 
    \textbf{Output:} $\mathcal{S}\coloneqq\{\mathcal{R}_k\}_{k=1}^{B}$ a set containing the bootstrap resamples of the empirical risk on the validation dataset.
    \begin{algorithmic}[1]
    \State $\Gamma\gets\texttt{fixedSequence}(\mathcal{D}_{\text{cal}},\Tilde{\Lambda},\alpha,\delta)$\textcolor{olive}{ \Comment{Calibrate using the fixed sequence procedure}}
    \State $\lambda\gets\text{max}(\Gamma)$\textcolor{olive}{ \Comment{Obtain a hyper-parameter from the calibration procedure}}
    \State $\mathcal{S}\gets  \varnothing$\textcolor{olive}{ \Comment{initialize the set to store the bootstrap resamples}}
    \For{$k\textbf{ in }1,\dots ,B$}
    \State  $I\gets \texttt{resample}(1,\dots, m)$ 
    \textcolor{olive}{\Comment{Obtain a resample from the indices  in the validation dataset \emph{with replacement}}}
    \State$\mathcal{R}_{k}\gets \frac{1}{m}\sum_{i\in I}\ell(\hat{Y}_{\text{full}}(X_{n+i}),\hat{Y}_{\lambda}(X_{n+i}))$
    \textcolor{olive}{\Comment{Evaluate the empirical risk}}
   \State $\mathcal{S}\gets \mathcal{S}\cup \{\mathcal{R}_{k}\}$
  \textcolor{olive}{ \Comment{$\mathcal{S}$ gets updated to include the generated risk}}
    \EndFor
   \\
    \Return $\mathcal{S}$
    \end{algorithmic}
\end{algorithm}

\begin{algorithm}[H]
    \caption{Fallback procedure to calibrate in the selective classification setting \cite{angelopoulos2022learn, wiens2005fallback}}\label{algorithm3}
    \hspace*{\algorithmicindent} \textbf{Inputs:} $\mathcal{D}_{\text{cal}}\coloneqq\{(X_i,Y_i)\}_{i=1}^{n}$,$\Tilde{\Lambda}=\Tilde{\Lambda}_1\times\Tilde{\Lambda}_2=\{\lambda_{1,k}:k\in \{1,\dots, J\} \}\times\{ \lambda_{2,j}|j\in \{0,\dots, T\}\}$,$\big\{\delta_{k,j}|k\in \{1,\dots, J \}\text{ and }j\in \{0,\dots,  T\}\big\}$ initial error allocations, $\alpha\in (0,1)$ our desired risk control threshold. \\
    \hspace*{\algorithmicindent} 
    \textbf{Output:} $\Gamma$ a set containing the $(\lambda_1, \lambda_2)$ pairs whose corresponding null hypotheses were rejected
    \begin{algorithmic}[1]
    \For{$k\textbf{ in }1,\dots ,J$}
    \For{$j\textbf{ in }0,\dots ,T$}
     \State  $\mathcal{I}_{\lambda_{1,k}, \lambda_{2,j}}\gets \{i\in \{1,\dots, n \}| \Hat{Y}_{\lambda_{2,j}}(X_i)>\lambda_{1,k}\}$ 
    \textcolor{olive}{\Comment{Compute the relevant indices .}}
    \State  $\hat{R}_{k,j}\gets \frac{1}{n_{\lambda_{1,k},\lambda_{2,j}}}\sum_{i\in \mathcal{I}_{\lambda_{1,k}, \lambda_{2,j}}} \mathbbm{1}\big\{ \hat{Y}_{\lambda_{2,j}}(X_i)\neq Y_{i}\big\}$ 
    \textcolor{olive}{\Comment{Compute the selective empirical risk.}}
    \State  $p_{k,j}\gets\mathbb{P}\big( Bin(n,\alpha)\leq n\Hat{R}_{k,j}\big) $ 
    \textcolor{olive}{\Comment{Compute a valid p-value.}}
    
   \If{$p_{k,j}\leq \delta_{k,j}$}
   \State $\Gamma \gets \Gamma\cup \{(\lambda_{1,k}, \lambda_{2,j})\}$
  \textcolor{olive}{\Comment{$H_{\lambda_{1,k},\lambda_{2,j}}$ gets rejected.}}
  \If{$j\leq T-1$}
  \State $\delta_{k,j+1}\gets \delta_{k,j+1}+\delta_{k,j}$
  \textcolor{olive}{\Comment{Distribute the budget allocation.}}
  \Else
  \If{$j== T \textbf{ and }k<J$}
    \State $\delta_{k+1,0}\gets \delta_{k+1,0}+\delta_{k,T}$
      \textcolor{olive}{\Comment{Distribute the budget allocation.}}
      \EndIf
      \EndIf
      \EndIf
    \EndFor
     \EndFor
   \\
    \Return $\Gamma$
    \end{algorithmic}
\end{algorithm}

\end{document}